\documentclass[10pt]{article}   
\usepackage[a4paper,margin=1.4in]{geometry}

\usepackage{pdfsync,amssymb,amsmath,ifpdf,color,placeins}

\newcommand{\R}{{\mathbb{R}}} 
\newcommand{\T}{{^\text{T}}}             %
\newcommand{\W}[1]{\mathbf{W}^\text{#1}} %
 
\newcommand{\n}[1]{{N_\text{#1}}} 
\def\u{\mathbf{u}} %
\newcommand{\x}{\mathbf{x}}
\newcommand{\teach}{\text{target}}
\newcommand{\y}{\mathbf{y}} 
\newcommand{\yt}{\mathbf{y}^\teach} 
\newcommand{\concat}[2]{{[#1;#2]}}
\newcommand{\X}{\mathbf{X}}
\newcommand{\Y}{\mathbf{Y}}
\newcommand{\Yt}{\mathbf{Y}^\teach}

\newcommand{\I}{\mathbf{I}}
\newcommand{\A}{\mathbf{A}}

\newcommand{\OO}{{\BigO}}

\usepackage{booktabs}%
\newcommand{\tableHeader}[1]{\textbf{#1}}

\usepackage{float}
\usepackage{multirow}
\usepackage{graphicx}

\usepackage[hang]{subfigure}
\usepackage{adjustbox} %
\usepackage{tabularx}
\usepackage{mdwtab} %

\newcommand{\BigO}{\mathcal{O}}

\definecolor{refcolor}{rgb}{0,0,0.5}
\ifpdf
\PassOptionsToPackage{hyphens}{url}
\usepackage[%
  pdftitle={Efficient implementations of echo state network cross-validation},%
  pdfauthor={Mantas Lukosevicius, Arnas Uselis},%
  pdfkeywords={Echo State Network, Reservoir Computing, Recurrent Neural Networks, Cross-Validation, Time-Complexity},%
  pdfstartview=FitH,%
  bookmarks=true,%
  bookmarksopen=true,%
  breaklinks=true,%
  colorlinks=true,%
  linkcolor=refcolor,%
  anchorcolor=blue,%
  citecolor=refcolor,%
  filecolor=blue,%
  menucolor=blue,%
  urlcolor=refcolor,%
  pdfpagelabels]{hyperref} 
\else 
\usepackage[%
  breaklinks=true,%
  colorlinks=true,%
  linkcolor=blue,%
  anchorcolor=blue,%
  citecolor=blue,%
  filecolor=blue,%
  menucolor=blue,%
  urlcolor=blue]{hyperref} 
\fi

\title{Efficient Implementations of Echo State Network Cross-Validation}

\author{Mantas Luko\v{s}evi\v{c}ius \href{https://orcid.org/0000-0001-7963-285X}{\includegraphics[scale=0.5]{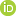}} \\ \texttt{mantas.lukosevicius@ktu.edu} \and
Arnas Uselis \\ \texttt{auselis@gmx.com}
}

\date{ Kaunas University of Technology, Kaunas, Lithuania   \\[2ex]%
        December 3, 2020}
				
\begin{document}

\maketitle              %

\begin{abstract} 

\textbf{Background/introduction:} Cross-Validation (CV) is still uncommon in time series modeling. Echo State Networks (ESNs), as a prime example of Reservoir Computing (RC) models, are known for their fast and precise one-shot learning, that often benefit from good hyper-parameter tuning. This makes them ideal to change the status quo.

\textbf{Methods:} We discuss CV of time series for predicting a concrete time interval of interest, suggest several schemes for cross-validating ESNs and introduce an efficient algorithm for implementing them. This algorithm is presented as two levels of optimizations of doing $k$-fold CV. Training an RC model typically consists of two stages: (i) running the reservoir with the data and (ii) computing the optimal readouts. The first level of our optimization addresses the most computationally expensive part (i) and makes it remain constant irrespective of $k$. It dramatically reduces reservoir computations in any type of RC system and is enough if $k$ is small. The second level of optimization also makes the (ii) part remain constant irrespective of large $k$, as long as the dimension of the output is low. We discuss when the proposed validation schemes for ESNs could be beneficial, three options for producing the final model and empirically investigate them on six different real-world datasets, as well as do empirical computation time experiments. We provide the code in an online repository.

\textbf{Results:} Proposed CV schemes give better and more stable test performance in all the six different real-world datasets, three task types. Empirical run times confirm our complexity analysis.

\textbf{Conclusions:} In most situations $k$-fold CV of ESNs and many other RC models can be done for virtually the same time and space complexity as a simple single-split validation. This enables CV to become a standard practice in RC.

\end{abstract}

\section{Introduction}

Echo State Network (ESN) \cite{Jaeger01a,JaegerHaas04,Jaeger07scholarpedia} is a recurrent neural network training technique of reservoir computing type \cite{LukoseviciusJaeger09}, known for its fast and precise one-shot learning of time series. But it often benefits from good hyper-parameter tuning to get the best performance. For this fast and representative validation is very important. 

Validation aims to estimate how well the trained model will perform in testing, which in turn is a proxy for the real-world performance. Typically ESNs and other time series modeling methods are validated on a single data subset. This single training-validation split is just one shot at estimating the test performance. Doing several splits and averaging the results can make a better estimation. This is known as cross-validation (\cite{Stone74} is often cited as one of the earliest descriptions), as the same data can be used for training in one split and for validation in another, and vice versa. 

Cross-validation is a standard technique in static non-temporal machine learning tasks, but is not that common in temporal modeling. For this we see two main reasons:
\begin{itemize}
    \item Splitting time series for training and validation can be problematic for some models, especially when a training part comes after a validation part. Temporal dependencies, information, and error gradient leaks across the cuts have to be disentangled. %
    \item Training of temporal models, in particular neural networks, is often a computationally expensive (not well parallelizable) and delicate procedure. Cross-validation further multiplies the computational cost and the probability of a failure several times.
\end{itemize}
Echo state networks have less problems in both regards than many other temporal modeling methods, in particular neural networks, and are relatively well-positioned for cross-validation ``out of the box'', but we make them even more so here.

In this contribution we suggest several schemes for cross-validating ESNs and introduce two stages of efficient algorithms for implementing them together with their time and space complexity analysis. In the first stage we optimize the use of the reservoir (any reservoir), which is the most expensive part. It makes reservoir running independent from the number of validation splits. The second stage addresses optimizing the output weight computations for all the different splits that can become most expensive when the number of splits becomes large. This second optimization is based on recomputing inverses of modified matrices \cite{Woodbury50,ShermanMorrison49} which is not that different to how it is done in the classical online learning Recursive Least Squares algorithm (used with ESNs early on \cite{Jaeger03}) or was independently applied to feed-forward models, e.g., \cite{ZhaoWang14,ShaoErWang16}.

We test the validation schemes on six different real-world datasets. We also run numeric simulations to empirically estimate the training times of the different algorithms. We share the code publicly at \url{https://github.com/oshapio/Efficient-Cross-Validation-of-Echo-State-Networks}.

The goal of the experiments here is not to obtain the best possible performance on the datasets, but to compare different validation methods of ESNs on equal terms. The best performance here is often sacrificed for the simpler models and procedures. We use classical ESNs here, but the proposed validation schemes apply to any type of reservoirs with the same time and space complexity savings, as long as they have the same linear readouts.

An earlier shorter version of this article was presented and published as a conference publication in \cite{LukoseviciusUselis19}. Compared to it, here we expand our argumentation and overview of related work, add the second level of the cross-validation algorithm optimization (computing outputs for large number of validation splits), repeat and average experiments over multiple random initializations, add experiments with a real-world medical time series output task, and add empirical computation time measurements. 

We introduce our ESN model, training, and notation in Section \ref{esnBasics}, discuss different types of tasks that might be important for validation in Section \ref{tasks}, discuss cross-validation nuances of time series in Section \ref{temporalXVal}, suggest several cross-validation schemes for ESNs in Section \ref{validations} and several ways of producing the final trained model in Section \ref{finalModel}. We introduce a time- and space-efficient algorithm for cross-validating ESNs: the efficient use of the reservoir in Section \ref{algorithm}, the efficient output weight computations for many folds in Section \ref{largek}, and summary in Section \ref{methods_summary}. We also report empirical experiments with different types of data in Section \ref{experiments}, empirical training time estimation in Section \ref{speed}, and conclude with a discussion in Section \ref{discussion}.

\subsection{Basic ESN Training}\label{esnBasics}

Here we introduce our ESN model and notation based on \cite{Lukosevicius12a}. 

The typical update equation of ESN is
\begin{equation}
\x(n)= (1-\alpha)\x(n-1)+\alpha \tanh \left( \W{in}\concat{1}{\u(n)}+\W{} \x(n-1) \right),
\label{eq:esnupdate} 
\end{equation} 
where $\x(n) \in \R^\n{x}$ is a vector of reservoir neuron activations and $\u(n) \in \R^\n{u}$ is the input, both at time step $n$; $\W{in}\in \R^{\n{x} \times (1+\n{u})}$ and $\W{} \in \R^{\n{x} \times \n{x}}$ are the input and recurrent weight matrices respectively; $\alpha \in (0,1]$ is the leaking rate; $\tanh(\cdot)$ is applied element-wise, and $\concat{\cdot}{\cdot}$ stands for a vertical vector (or matrix) concatenation. 

A typical linear readout layer is
\begin{equation}
\y(n)=\W{out}[1;\u(n);\x(n)],
\label{eq:linreadout}
\end{equation}
where $\y(n) \in \R^\n{y}$ is the network output at time step $n$ and $\W{out} \in \R^{\n{y}\times\n{r}}$ is the output weight matrix. We denote $\n{r} = 1+\n{u}+\n{x}$ as the size of the ``extended'' reservoir $[1;\u(n);\x(n)]$ for brevity.

Equation \eqref{eq:linreadout} can be written in a matrix form as
\begin{equation} 
\Y = \W{out}\X, 
\label{eq:readoutM} 
\end{equation}
where $\Y \in \R^{\n{y} \times T}$ are all $\y(n)$ and $\X \in \R^{\n{r} \times T}$ are all $[1;\u(n);\x(n)]$ obtained by presenting the reservoir with $\u(n)$, both collected into respective matrices by concatenating the column-vectors horizontally over the training period $n=1,\ldots,T$. %

Finding the optimal weights $\W{out}$ that minimize the squared error between $\y(n)$ and $\yt(n)$ amounts to solving a system of linear equations 
\begin{equation} 
\Yt = \W{out}\X 
\label{eq:regressTargetM} 
\end{equation}
with respect to $\W{out}$ in a least-square sense, i.e., a case of linear regression. 
The system is typically overdetermined because $T \gg \n{r}$.
$\Yt \in \R^{\n{y} \times T}$ here are all $\yt(n)$ collected similarly to $\X$.

The most commonly used solution to \eqref{eq:regressTargetM} in this context is ridge regression
\begin{equation} 
\W{out}=\Yt \X^\T \left(\X\X^\T + \beta \I \right)^{-1},
\label{eq:ridgeRegress} 
\end{equation} 
where $\beta$ is a regularization coefficient and $\I$ is the identity matrix. %
It is advisable to set the first element of $\I$ in \eqref{eq:ridgeRegress} to zero to exclude the bias connection from the regularization. 

For more details on generating and training ESNs see \cite{Lukosevicius12a}.

\section{Validation of Echo State Networks}\label{proposedMethods}

In this section we discuss several validation options for ESNs, and propose an efficient algorithm for their cross-validation.

\subsection{Different Tasks}\label{tasks}

Some details of validation implementation and optimization depend on the type of the task we are learning. Let us differentiate temporal machine learning into three types of tasks:
\begin{enumerate} 
    \item \textbf{Generative} tasks, a closed-loop scenario, where the computed output $\y(n)$ comes back as (part of) the input $\u(n+m)$.\footnote{Note that this can alternatively be implemented by feedback connections $\W{fb}$ from $\y(n-m)$ to $\x(n)$ in \eqref{eq:esnupdate} \cite{Lukosevicius12a}.} These tasks can be pattern generation or multi-step time series prediction in a generative mode.
    \item \textbf{Output} tasks, an open-loop scenario, where the computed output $\y(n)$ does not come back as part of the input. These tasks can be detection or recognition in time series, deducing a signal from other contemporary signals. 
    \item \textbf{Classification} tasks, of separate (pre-cut) finite sequences, where a class $\y$ is assigned to each whole sequence $\u(n)$. 
\end{enumerate}

For the classification tasks we usually store only an averaged state or a concatenation of fixed number of states $\x(n)$ for every sequence in the state matrix $\X$ \cite{Lukosevicius12a}. This type of tasks is basically the non-temporal classification, after the states have been produced.

Our experiments Section \ref{experiments} includes tasks of all the three types and is structured accordingly. 

\subsection{Cross-Validation in Time Series}\label{temporalXVal}

$k$-fold cross-validation, arguably the most popular cross-validation type, is a standard technique in static (non-temporal) machine learning tasks where data points are independent of each other \cite{Arlot2010}. Using this scheme, the data are partitioned into $k$ usually equal folds, and $k$ different train-validate splits of the data are done, where one (each time different) fold is used for validation and the rest for training.

Temporal data, on the other hand, are time series or signals, often a single continuous one. They are position- or sequence-dependent by definition. This breaks the sample independence assumption and makes the theoretical framework for cross-validation in time series more challenging. This is one of the reasons why cross-validation in them is less common, but it is gaining some recent interest nonetheless \cite{CerqueiraEtAl2020}.

In this contribution, we are mostly concerned with producing the model that can best predict the data directly following the one we have for training and validation. This is a common real-world scenario when we want to peek into the future. This is in contrast to estimating the best unbiased model of the data in a some more abstract sense. Here we make no assumption that the generating process is strictly stationary.

Simulating and evaluating this prediction into the future we will call testing, as opposed to validation, and do this identically for all the validation schemes tested.

Statically splitting the data into initialization, training, validation, and testing, in that order in time, is the most common way to train, validate, and test ESNs, as well as many other temporal models in such a scenario. This validation scheme is often simply referred to as out-of-sample validation. The short initialization (also called transient) phase is used to get the state of the reservoir $\x(n)$ ``in tune'' with the input $\u(n)$ \cite{Lukosevicius12a}. This sequence is finite, and often quite short, because ESNs possess the echo state property \cite{YildizEtAl12}. Initialization is only necessary at the start, because the subsequent phases can take the last $\x(n)$ from the previous phase if data continue without gaps. When training for generative tasks the real (future) outputs $\y(n)$ are substituted with targets $\yt(n)$ in inputs, known as ``teacher forcing'', to break the cyclic dependency. 

Because the memory of ESN is preserved in its state $\x(n)$, and the classical readout that is learned is memory-less, we can do the instant switches between the phases. Since there are no errors back-propagated through time, no target $\yt(n+m)$ information is leaked from the future $n+m$. The past targets $\yt(n-m)$ generally have also no influence on learning $\y(n)$. This Markovian property of the readout, makes cross-validation with ESNs rather straightforward. The only information that leaks across the fold cuts is the past inputs $\u(n-m)$, that leave their traces in $\x(n)$. 

Leaving temporal gaps between training and validation folds is often suggested to minimize the dependencies \cite{racine2000consistent,Arlot2010,CerqueiraEtAl2020,BergmeirEtAl18}. We include them as variations in most of our experiments, but also have arguments against them:
\begin{itemize}
    \item Less data are used for training and validation.
    \item If the data are non-stationary the gaps would not get rid of the dependencies anyway.
    \item For our particular setup, where the data are not stationary and we want to predict the data that directly follows, we want to use the data for training/validation up to the last point before the prediction, as it is likely the most similar to the testing one and thus the most valuable. And we want our validation to emulate this testing, thus no gaps between training and validation either.
\end{itemize}

In other words, we embrace the non-stationarity and try to bias our model for a particular segment of the time series. The same is in fact done by the above-mentioned classical train-validate-test sequential scheme. Putting more emphasis on the ending of the training data that is closer to the testing is also possible with ESNs by a time-
weighting scheme \cite{Lukosevicius12a} and helps in practice \cite{DaukantasEtal10}.

Another likely reason for why cross-validation is not popular in temporal models is that training them is typically computationally expensive, and, in the case of fully-trained recurrent neural networks, a potentially unstable process \cite{Doya92,BengioEtal94,PascanuEtAl13}. Doing a $k$-fold cross-validation multiplies these difficulties $k$ times. ESNs are relatively well-suited for cross-validation in this respect as they are, but we make them even more so in this work.

We see the following intuitive cases when using cross-validation with time series could be beneficial:
\begin{itemize}
    \item When the data are scarce, cross-validation efficiently uses all of the available data for both training and validation.
    \item Combining the models trained on different folds could be a form of (additional) regularization, improving stability. 
    \item A similar argument holds when the data have some occasional anomalies or imperfections: averaging validation over many folds reduces their effects.
    \item Generally speaking, when the process generating the data is not strictly stationary and it ``wanders'' around (back and forth), cross-validation increases the chances that the testing interval is adequately covered by the model. 
\end{itemize}

However, if the process ``drifts'' in one direction, validating and tuning the hyper-parameters on the data interval directly following the training interval might be beneficial; and picking the validation interval adjacent to the testing one (i.e., the classic single validation) might in fact be the best option.

In particular, we do not expect a validation scheme to make a difference on long stationary time series, like (synthetic) chaotic attractors, as it does not matter which (and to some extent how long, if the data are ample) sections of the data are taken for training or validation.

\subsection{Validation Schemes}\label{validations}

Here we suggest several cross-validation schemes for ESNs.

\begin{figure}[htb]%
    \centering
    \includegraphics[width=.75\textwidth]{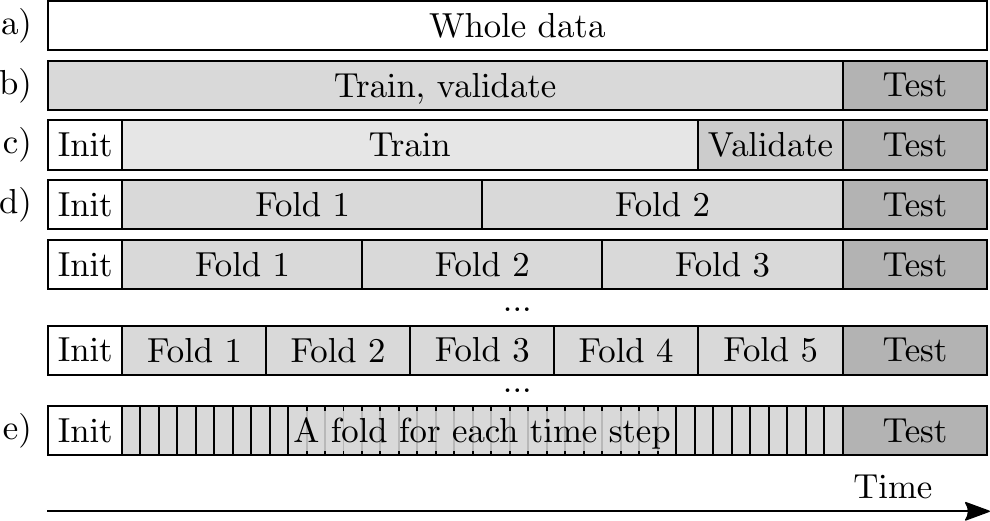} 
        \caption{Splitting the data: a) all the available data; b) splitting-off the testing set; c) the classical Single Validation (SV) split for ESNs; d) splitting the data into 2 folds and up for $k$-fold cross-validation; e) the maximum amount of folds for leave-one-out cross-validation.}
    \label{fig:data_splits}%
\end{figure}

We firstly separate the testing part off the end of the data, which is independent of the validation scheme and is left for testing it as illustrated in Figure \ref{fig:data_splits}b. 

The classical ESN validation scheme explained above is depicted in Figure \ref{fig:data_splits}c. We will refer to it as Single Validation (SV). %

For $k$-fold cross-validation of ESNs we split the data into $k$ folds. The number of folds $k$ can be varied from 2, as shown in Figure \ref{fig:data_splits}d, up to available data/time points ending up with leave-one-out cross-validation \ref{fig:data_splits}e.

In addition to the classical SV split we investigate these iterative validation schemes of using the data between the initialization and testing parts:
\begin{enumerate}

    \item \textbf{$k$-fold Cross-Validation} (CV). The data are split into $k$ equal folds. Training and validation are performed $k$ times (doing $k$ training-validation splits), each time taking a different single fold for validation and all the rest for training.  

    \item \textbf{$k$-fold Accumulative Validation} (AV). First, we split the ``minimum'' required amount for training only off the beginning, then we divide the rest of the data into $k$ equal folds. Training and validation are performed $k$ times, each time validating on a different fold, similarly to CV, but only training on all the data preceding the validation fold. 
    
    \item \textbf{$k$-fold walk-Forward Validation} (FV) is similar to AV: the splitting is identical and validation is done on the same folds, but the training is each time done only on the same fixed ``minimal'' amount of data directly preceding the validation fold. 

\end{enumerate}

\begin{figure}[htb]%
    \centering
    \includegraphics[width=\textwidth]{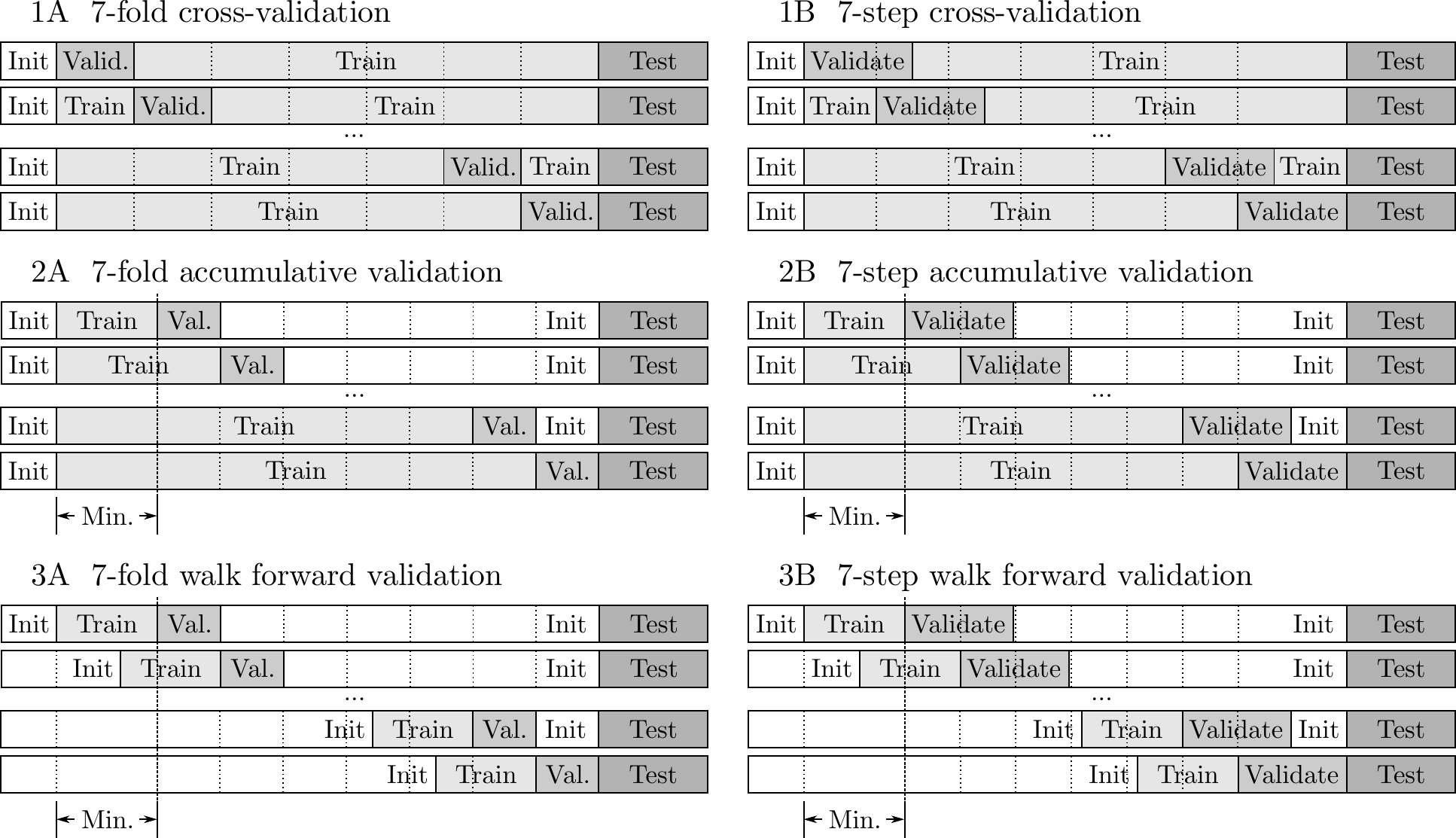} 
        \caption{Different validation schemes investigated.}
    \label{fig:validation_types}%
\end{figure}

The three validation schemes are illustrated in the left column of Figure \ref{fig:validation_types}. 

Each validation scheme has some rationale behind it. 

In CV all the data are used for either training or validation in each split. Also, all the data are used exactly $k-1$ times for training and 1 time for validation. This has its benefits explained later in Section \ref{finalModel}. What is a bit unusual for temporal data here is that the training data can come later in the sequence than the validation data. But this should not necessarily be considered a problem, as discussed in Section \ref{temporalXVal} and in \cite{BergmeirEtAl18}. In fact, for time series output and classification tasks the columns of $\X$ and $\Yt$ could in principle be randomly shuffled, before applying $k$-fold cross-validation, as is common in non-temporal tasks. 

CV is geared towards training the model once on a fixed representative dataset, that are perhaps scarce. It does not, however, try to capture the temporal drift of the generating process and bias the model for the interval directly after the training/validation. It is perhaps more suitable for a stationary process or where the nonstationarity is abrupt and hard to extrapolate. 

AV emulates the classical single training and validation SV $k$ times by each time training on all the available data that come before the validation fold and then validating. This scenario would also happen when the model is repeatedly updated using SV with continuously arriving new data. This idea is not new, of course \cite{Dawid1984}. Here we do not allow our model to ``peek into the future''. The downsides are that models are not trained on the same amount of data and more models are trained on the beginning of the data more than the ending. 

FV is similar to AV but keeps the training data length constant like in CV, which is more consistent when selecting good hyper-parameters. Training length can be set to match several folds, which is quite common when doing walk-forward validation. Here the assumption on stationarity of the generating process is weaker, models are trained and validated ``locally'' in time, in a sliding window fashion. 

CV, AV, and FV all use progressively less data for training, thus, in general, are progressively less advisable when data are scarce. 

We also investigate the counterparts for the three validation schemes with validation ``folds'' set constant, independent of $k$. They are illustrated in the right column of Figure \ref{fig:validation_types}. We call them ``$k$-step'' as opposed to ``$k$-fold'', as here we are moving with a constant step instead of using the same nonoverlapping folds. We are defining the terminology here where we could not find a consistent one in the literature. 

These ``$k$-step'' validation schemes are mostly relevant to realistically validating generative tasks (as defined in Section \ref{tasks}), where due to the feedback loop, errors tend to escalate over time, sometimes resulting in the fast divergence of the outputs from the desired ones. Thus the length of the validation interval here is critically important and should be set to match the testing length of a realistic use case scenario (or perhaps a bit longer to have a margin of stability). For other types of tasks, where we simply average errors of independently computed outputs for every time step, the validation interval length is usually not that important.

The ``$k$-step'' validation schemes can introduce overlaps (like in Figure \ref{fig:validation_types} right) or gaps between the validation intervals, making the data points unequally represented in training and validation splits. This can be resolved by setting the validation and step lengths such, that validation matches an integer number of steps, and thus data points appear in the same number of validation intervals (excluding the ends). Setting validation fold length equal to the length of one step falls back to the ``$k$-fold'' versions. 

However, the bias would still not be gone completely, because in the generative mode not all the data samples in the validation interval are equal, as the interval is compared with a typically diverging output. This is true for both the ``$k$-step'' and ``$k$-fold''. The bias can be mostly eliminated in the limit case of ``$k$-step'', when step size is just one data sample, like in leave-one-out. This would be computationally costly despite our optimizations, as you still need to produce the generative output over the whole validation interval independently for each split, but well compatible with the ``$k$-step'' schemes. In our experiments we do not go to such extremes and consider the schemes where multiple validation splits are averaged being still less biased than the standard single SV validation. But one should perhaps look out for an interference between the step size and the periodicity of the series. 

We also do experiments were we introduce \textbf{gaps} between the training and validation intervals. For SV, AV, and FV we introduce the gap between the training and the validation intervals, and for VC we try gaps before the validation interval, after, and both. The gap sizes were equal to the validation interval size. Testing was done the same way as for the versions with no gaps to have it equal.

\subsection{The Final Trained Model}\label{finalModel}

The best hyper-parameter set is usually selected based on the average of validation results over the $k$ training-validation splits. We investigate several ways of producing the final trained model with the best hyper-parameters for testing:
\begin{itemize}
    \item \textbf{Retrain} ESN on all the training and validation data;
    \item \textbf{Average} $k$ $\W{out}$s of ESNs that have been trained on the $k$ splits; %
    \item Select the ESN that validated \textbf{best} among the $k$.
\end{itemize}

Each method again has its own rationale. Retraining uses all the available data to train an ESN in a straightforward way. However, this requires some additional computation and the hyper-parameters might no longer be optimal for the longer training sequence. Averaging $\W{out}$s introduces additional regularization, which adds to the stability of outputs. This method might make less sense with AV, since models are trained on different amounts of data. The best-validated split, on the other hand, was likely trained on the hardest parts of the data (and/or validated on the easiest). A weighting scheme among the $k$ splits could also be introduced when combining the models, as well as time step weighting as discussed in \cite{Lukosevicius12a}. 

\textbf{Regularization} parameter can be grid-searched for every split individually and efficiently \cite{Lukosevicius12a}, as opposed to other hyper-parameters, that are searched in the optimization loops outside the validation scheme. When averaging or choosing the best, we use ESNs with their best regularizations for that particular fold; whereas when retraining the regularization that is best on average (over the folds) is utilized. It could also be averaged and/or corrected for the training length. Regularization is most important in the generative tasks.

For the classic single split SV we also have two options: to either retrain the model on the whole data or to use the validated model as it is (equivalent to the other two methods).

\subsection{Efficient Reservoir Use}\label{algorithm}

Under the typical assumptions $T \gg \n{r}$, $\n{r} \gg \n{y}$, $\n{r} \gg \n{u}$, running the ESN reservoir \eqref{eq:esnupdate} is dominated by $\W{}\x$ which takes $\OO(\n{x}^2)$ operations per time step with dense $\W{}$, or $\OO(\n{x}^2 T)$ for all the data. This can be pushed down to $\OO(\n{x} T)$ with sparse $\W{}$ \cite{Lukosevicius12a} which is the same $\OO(\n{r} T)$ required to collect $\X$. To compute $\W{out}$ \eqref{eq:ridgeRegress}, collecting $\X\X^\T$ takes $\OO(\n{r}^2 T)$ and the matrix inversion takes $\OO(\n{r}^3)$ operations in practical implementations. Thus the whole training of ESN (dominated by collecting $\X\X^\T$) is back to
\begin{equation}
\OO(\n{r}^2 T + \n{r}^3 ) = \OO(\n{r}^2 T). 
\label{eq:Oesn} 
\end{equation}

Training dominates, as validating ESN has the same time complexity as simply running it. Thus time complexity of doing a straightforward ESN $k$-fold cross-validation is 
\begin{equation}
\OO(k\n{r}^2 T + k\n{r}^3 ) = \OO(k \n{r}^2 T).
\label{eq:Onaive} 
\end{equation} 

Space complexity can be pushed from $\OO(\n{r} T)$ for $\X$ down to $\OO(\n{r}^2)$ when collecting $\X\X^\T$ and $\Yt \X^\T$ on the fly, which also allows ESNs to be one-shot-trained on virtually infinite time sequences \cite{Lukosevicius12a}.

However, we do not need to rerun the ESN for every split. We can collect and store the matrices $\X\X^\T$ and $\Yt \X^\T$ for the whole sequence once. Then for every split we only run the reservoir on the validation fold. Validation folds should be arranged consecutively like in Figure \ref{fig:validation_types}.1A, so that after running one validation fold we can save the reservoir state $\x(n)$ to continue running the next validation fold of the next split. We collect $\X_i\X_i^\T$ and $\Yt_i \X_i^\T$ on the validation fold $i$ and subtract them from the global ones to compute $\W{out}_i$ for the particular split $i$:
\begin{equation} 
\W{out}_i=\left(\Yt \X^\T - \Yt_i \X_i^\T \right) \left(\X\X^\T - \X_i\X_i^\T + \beta \I \right)^{-1}.
\label{eq:ridgeRegressFold} 
\end{equation} 
If we are doing output or classification task (as in Section \ref{tasks}) and we can afford to store $\X_i$ of the validation fold $i$ in memory, we can reuse it to compute the validation output $\Y_i$ \eqref{eq:readoutM} after we got $\W{out}_i$. If not, we need to rerun the validation fold one more time for this. \textit{This way the ESN is run through the whole data only two or three times irrespective of $k$.}

Notice also, that the space complexity of the implementation with three runs remains $\OO(\n{r}^2)$. It is $\OO(\n{r}^2+\n{r}T/k)$ if we store $\X_i$. We could alternatively also store $\X_i\X_i^\T$ and $\Yt_i \X_i^\T$ for every fold $i$ and save one running through the data this way, by having space complexity $\OO(k\n{r}^2)$. 

The proposed algorithm pushes down the time complexity of preparing the $\X\X^\T$
matrices in $k$-fold cross-validation from $\OO(k \n{r}^2 T)$, that dominates in \eqref{eq:Onaive}, to $\OO(\n{r}^2 T)$. Adding the matrix inversions \eqref{eq:ridgeRegress}, that are now not necessarily dominated, our proposed more efficient algorithm for $k$-fold cross-validating ESNs has time complexity
\begin{equation}
\OO(\n{r}^2 T + k \n{r}^3).
\label{eq:Ogood} 
\end{equation} 

Thus we get a $k$ or $T/\n{r}$ time complexity speedup in a more efficient implementation \eqref{eq:Ogood} compared to naive \eqref{eq:Onaive}, depending on which multiplier is smaller. 

When the data sample length $T$ is many times larger than the ESN size $\n{r}$ (a typical case and a one where optimization is most relevant) and $k$ not too large, such that $k<T/\n{r}$, we can say that \textit{the proposed efficient algorithm permits doing ESN $k$-fold cross-validation with the same time complexity as a simple one-shot validation SV. The space complexity can also remain the same.}

\subsection{Efficient Matrix Inverses for Many Folds}\label{largek}

The efficient implementations described so far work best when the number of folds $k$ is relatively small. If the task at hand benefits from large $k$ (up to leave-one-out validation $k=T$), the total number of computations inevitably increases. It gets dominated by the $k$ matrix inversions that are needed to find $\W{out}_i$ \eqref{eq:ridgeRegress} for each fold $i$: the $\OO(k \n{r}^3)$ in \eqref{eq:Ogood}. 

In such a case this bottleneck can be further optimized. 

The inverse that we need to compute for each validation fold $i$ (of $k$) is the $(\X\X^\T + \beta \I - \X_i\X_i^\T)^{-1}$ in \eqref{eq:ridgeRegressFold}. %
Let us for the sake %
of brevity denote $\A \equiv \X\X^\T+\beta\I$. Applying Woodbury matrix identity \cite{Woodbury50} to this problem we get
\begin{equation}
    \left(\A - \X_i\X_i^\T \right)^{-1} = \A^{-1} + \A^{-1}\X_i\left( \I-\X_i^T \A^{-1}\X_i \right)^{-1}  \left( \X_i^\T\A^{-1} \right).
    \label{eq:Woodbury}
\end{equation}
In the case of leave-one-out validation, $\X_i$ becomes a vector and we can apply Sherman–Morrison formula \cite{ShermanMorrison49} as in \cite{ShaoErWang16} which is a special case of Woodbury. The same idea is used in a classical Recursive Least Squares algorithm where the optimal weights are computed online, see \cite{Jaeger03} as an example with ESNs.

Let us analyze the time complexity of this approach \eqref{eq:Woodbury}. We can obviously precompute $\A^{-1}$ once in addition to $\A$ at a one-time cost $\OO(\n{r}^3)$. $\X_i \in \R^{\n{r} \times T/k}$, thus all matrices have dimensions $\n{r}$ and $T/k$ in \eqref{eq:Woodbury}. We assume large $k$ here and thus $T/k < \n{r}$. The matrix inverted here is of size $T/k \times T/k$, which takes $\OO((T/k)^3)$ and is only better than the regular $\OO(\n{r}^3)$ if this assumption holds. The matrices outside the inversion should be multiplied together first in \eqref{eq:Woodbury} for an efficient implementation. These multiplications take $\OO(\n{r}^2 T/k)$ and so do the multiplications inside the inverse. The multiplication of the inverted matrix with the rest takes $\OO(\n{r} (T/k)^2)$. Under the $T/k < \n{r}$ assumption, the dominating term of the three is the $\OO(\n{r}^2 T/k)$. The bigger matrix addition of the matrices takes $\OO(\n{r}^2)$, but is not bigger than $\OO(\n{r}^2 T/k)$, since obviously $k \leq T$. 

Lastly, multiplying the inverted matrix in \eqref{eq:ridgeRegress} takes $\OO(\n{r}^2\n{y})$. Thus, the total time-complexity of cross-validating ESN by this method \eqref{eq:Woodbury} is 
\begin{equation}
    \OO\left(\n{r}^2 T + \n{r}^3 + k (\n{r}^2 T/k + \n{y}\n{r}^2)\right)=\OO(\n{r}^2 T + k\n{r}^2\n{y}), 
    \label{eq:Obigk}
\end{equation}
as opposed to \eqref{eq:Ogood}, under the assumptions $T/k < \n{r}$ and the previous $T > \n{r}$.

The space complexity remains the same as that of the previous methods.

\textit{In case the output is one-dimensional $\n{y} = 1$ (or $\n{y} < T/k$), we have an ESN cross-validation algorithm with time complexity $\OO(\n{r}^2 T)$: neither time nor space complexity depends on the number of folds $k$, no matter how large, and remain the same as for single validation SV}. 

\subsection{Efficient Algorithms Summary}\label{methods_summary}

The best complexities of the different algorithms are summarized in Table \ref{tab:methods}. Efficient ESN implementations have already been discussed in \cite{Lukosevicius12a}. Here we argue for ESN cross-validation and propose efficient algorithms for that. A naive straightforward $k$-fold cross-validation would take $k$ times as long as doing a single training and validation. The time complexity is dominated by running the reservoir $k$ times. In Section \ref{algorithm} we have proposed an optimization, which allows to run the reservoir up to three times (for the most space-frugal version), irrespective of $k$. If $k$ is relatively small ($k<T/\n{r}$), the time complexity of this algorithm is virtually the same as not doing cross-validation at all: the dominating term $\OO(k \n{r}^2 T)$ is pushed back to $\OO(\n{r}^2 T)$. We call this algorithm ``Small $k$'' in Table \ref{tab:methods}. 

\begin{table}[h]                           
 \centering
    \caption{Summary of best complexities of the different cross-validation algorithms}
    \begin{tabular}{ l r r}
    \toprule
     \tableHeader{Algorithm} & \tableHeader{Time complexity} & \tableHeader{Space complexity} \\
    \midrule
    No cross-validation (SV) & $\OO(\n{r}^2 T)$      & $\OO(\n{r}^2)$   \\ 
    Naive                    & $\OO(k \n{r}^2 T)$    & $\OO(\n{r}^2)$   \\ 
    ``Small $k$''            & $\OO(\n{r}^2 T + k\n{r}^3)$ & $\OO(\n{r}^2)$ \\ 
    ``Large $k$''            & $\OO(\n{r}^2 T + k\n{r}^2\n{y})$ & $\OO(\n{r}^2)$ \\ 
    \bottomrule 
    \end{tabular}
    \label{tab:methods}                            
\end{table}

Each time the complexity-dominating term is sufficiently optimized, a new potentially dominating term emerges that has been dominated before. The ``Small $k$'' algorithm can get dominated by the $k \n{r}^3$ term when $k$ increases significantly. In Section \ref{largek} we discuss a method that pushes this $k \n{r}^3$ term back to the same $\n{r}^2T$, assuming $k$ is sufficiently large: $k > T/\n{r}$ so that the folds are smaller than the extended reservoir $T/k < \n{r}$. We call this algorithm ``Large $k$'' in Table \ref{tab:methods}. Even for $k$ approaching its maximum $T$ this algorithm gives virtually the same complexity as no cross-validation, provided that output dimension $\n{y}$ is small. The $k \n{r}^2\n{y}$ is the new potential term that emerges after all the bigger ones are optimized. 

The proposed algorithms push time complexity of ESN cross-validation back to $\OO(\n{r}^2 T)$ of non-cross-validated ESN training, which is the best we can hope for, for many realistic situations. 

All the different methods can be implemented to have the same original space complexity $\OO(\n{r}^2)$. This space frugality can be sacrificed to have a better practical speed and further save some reservoir running, even though the time complexity remains the same.

We have outlined an efficient method for ESN $k$-fold cross-validation (CV) here, but it can easily be adapted to other types of validation schemes described in Section \ref{validations}.

\section{Precision Experiments}\label{experiments}

Having established that different validation schemes for ESNs are possible and can be implemented efficiently, in this section we test their usefulness empirically on several different time series datasets. We run experiments with all the three task types discussed in Section \ref{tasks}, that are presented in the following subsections: generative, time series output, and classification. 

As mentioned before, the goal here is not to strive for the best possible performance but to compare different validation methods of simple ESNs on equal terms.

\subsection{Generative Mode}\label{generativeExperiments}

We evaluate the proposed validation methods by multi-step predicting four classical univariate time series datasets of increasing sizes in a generative mode:
\begin{itemize}
    \item \textbf{Labor}: ``Monthly unemployment rate in US from 1948 to 1977''\footnote{Publicly available at \url{https://data.bls.gov/timeseries/lns14000000}};
    \item \textbf{Gasoline}: Weekly ``US finished motor gasoline product supplied''\footnote{Publicly available at \url{https://www.eia.gov/dnav/pet/hist/LeafHandler.ashx?n=PET&s=wgfupus2&f=W}};
    \item \textbf{Sunspots}: ``Monthly numbers of sunspots, as from the World Data Center''\footnote{Publicly available at \url{http://www.sidc.be/silso/datafiles}};
    \item \textbf{Electricity}: ``Half-hourly electricity demand in England''\cite{Taylor2003}.

\end{itemize}

Lengths of the datasets together with testing and validation split parameters for different schemes are presented in Table \ref{tab:generative_datasets}. ``Min. ratio'' here is the percentage of the whole data (excluding testing) used as the minimal training length in AV or the whole in FV (the ``Min.'' in Figure \ref{fig:validation_types}). 

\begin{table}[h]                           
 \centering
    \caption{Datasets and validation setup parameters. }
    \begin{tabular*}{\textwidth}{@{\extracolsep{\fill}} l r r r r}
    \toprule
     \tableHeader{Dataset} & \tableHeader{Samples $T$} & \tableHeader{Valid, test samples} & \tableHeader{Folds, steps $k$} & \tableHeader{Min. ratio}\\
    \midrule
     Labor & 360 & 10 & 34 & 50\,\% \\
     Gasoline & 1\,355 & 67 & 18 & 50\,\% \\ 
      Sunspots & 3\,177 & 200 & 10 & 50\,\% \\ 
      Electricity & 4\,033 & 200 & 18 & 50\,\% \\ 
     \bottomrule  
    \end{tabular*}
    \label{tab:generative_datasets}                            
\end{table}

As discussed, for generative tasks the length of the validation fold is of crucial importance and thus we use the ``$k$-step'' validations. Here we use validation length the same as testing to have comparable conditions and error measures, but it could maybe be even longer for an additional stability margin. For $k$-step CV, the initialization length and $k$ are chosen such that the $k$ folds would all have the same length as testing, thus $k$-step CV becomes $k$-fold CV. For the other two $k$-step validation schemes we use the same validation length. %

\begin{figure}[h]
 \centering
 \includegraphics[width=\textwidth]{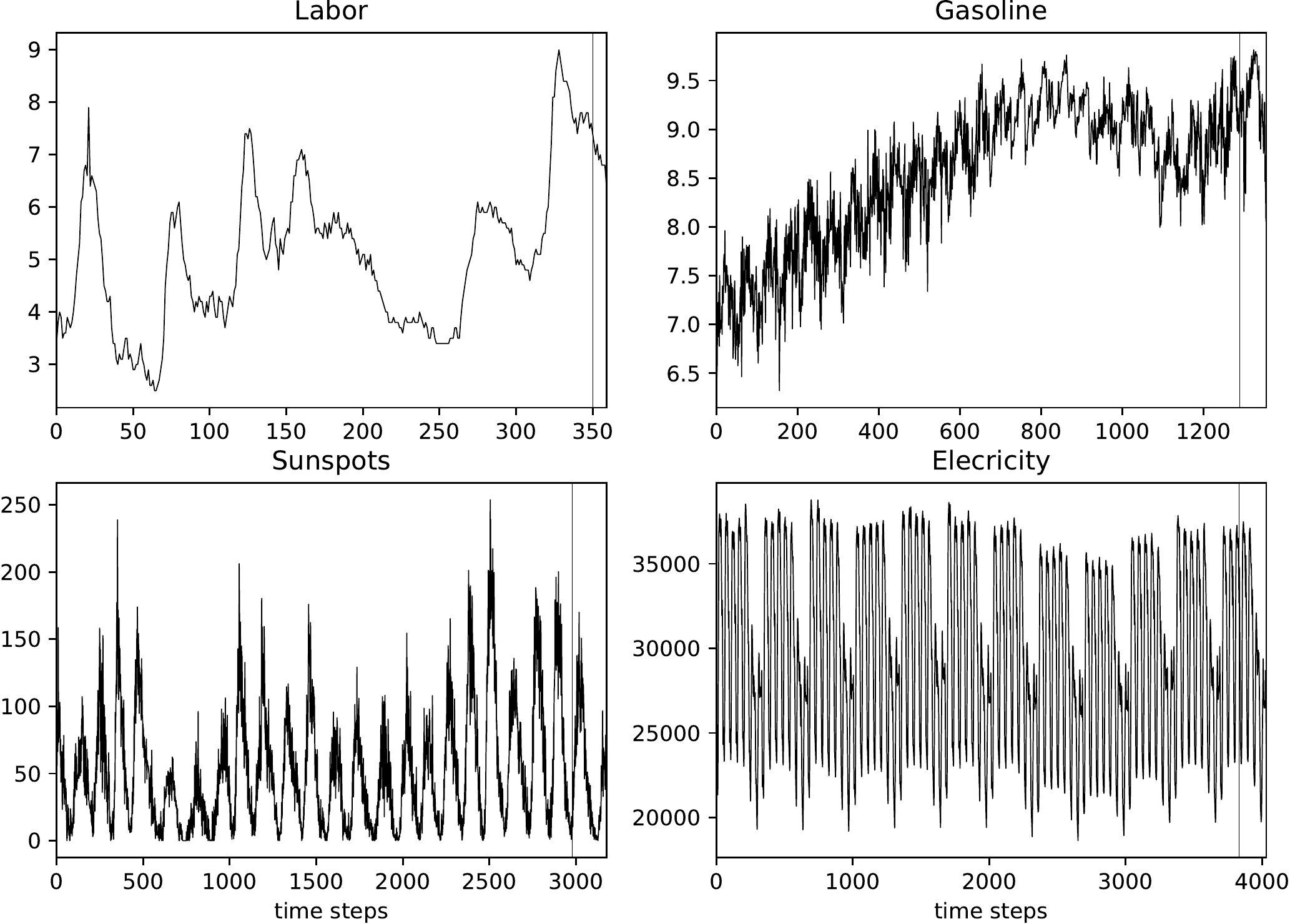}
\caption{Time series datasets used for prediction in generative mode. The testing intervals are separated at the end.}
 \label{fig:serieses}
\end{figure}

The datasets are plotted in Figure \ref{fig:serieses} with testing intervals indicated at the end of each. We see that the Gasoline and Electricity have strong seasonal components as well as trends. For the most accurate predictions these should perhaps be separated, making some components more stationary; but, as mentioned, here we are not striving for this, and do not want to hide from the nonstationarity of the data, that can not be all removed this way. The Sunspots dataset, while seems to also have a strong periodic component, is more likely chaotic and almost impossible to forecast accurately over the period of 24 to 48 monthly time steps \cite{rozelot1995chaotic}. The Labor series, however, looks the hardest to predict, because it is the scarcest, the least regular, and takes a sudden dive in the testing interval.

We use a grid search to find the best hyper-parameters of ESN. The reservoir size was fixed at $\n{x} = 50$. The best leaking rate $\alpha \in \{ 0.1, 0.2, 0.3,$ $\ldots, 1\}$, spectral radius $\rho \in \{0.1, 0.2, 0.3, \ldots, 1.5\}$, and regularization degree $\beta \in \{ 0, 10^{-9}, 10^{-8}, \ldots, 1 \}$ were searched in the respective sets following \cite{Lukosevicius12a}, each parameter combination validated using the proposed schemes. This procedure was carried out five times with different random ESN weight $\W{in}$ and $\W{}$ initializations and statistics of the results computed.

The experiment results are presented in Table \ref{tab:generative_results}. The results are with the investigated validation schemes, with and without the gaps around the validation fold, and different ways of producing the final trained model. We do not present standard deviations here for visual clarity, but the dispersion of the testing errors among the variations can give some indication of it. 

\bgroup
\setlength\tabcolsep{0pt}
\begin{table}[H]                           
 \centering
    \caption{Average validation and testing NRMSEs on generative datasets over five initializations with the different validation schemes, without and with gaps before (\_V), after (V\_), or both sides (\_V\_) the validation fold. Several best testing values for every dataset are highlighted in bold.}
        \begin{tabular*}{\textwidth}{ @{\extracolsep{\fill}} l l r r r r r r r r }
\toprule
     \multicolumn{2}{c} {\tableHeader{Method}} & \multicolumn{2}{c}{\tableHeader{Labor}}  &  \multicolumn{2}{c}{\tableHeader{Gasoline}} &       \multicolumn{2}{c}{\tableHeader{Sunspots}} & \multicolumn{2}{c}{\tableHeader{Electricity} } \\
\cmidrule(r){1-2} \cmidrule(lr){3-4} \cmidrule(lr){5-6} \cmidrule(lr){7-8} \cmidrule(l){9-10}  
    \multicolumn{1}{l} {\tableHeader{Validation}} &    \multicolumn{1}{l} {\tableHeader{Final}} & \textbf{Valid} & \textbf{Test} & \textbf{Valid}& \textbf{Test}& \textbf{Valid}& \textbf{Test} & \textbf{Valid}& \textbf{Test} \\
\midrule
     \multirow{2}{*}{SV} & As is & \multirow{2}{*}{0.957} & 1.504  & \multirow{2}{*}{0.875} & 1523.949 & \multirow{2}{*}{0.673} & 0.915 &  \multirow{2}{*}{0.358} &2902.103   \\ 
     &  Retrained &     & \textbf{1.284}   &  & 120.525  &   & 0.877  &  & 4.679  \\ 
	[3pt]\multicolumn{1}{r}{\multirow{2}{*}{\_V}} & As is & \multirow{2}{*}{1.016} & 2.181  & \multirow{2}{*}{0.874} & 1620.997 & \multirow{2}{*}{0.639} & 1.063 &  \multirow{2}{*}{0.353} & 0.736  \\ 
     &  Retrained &     & 1.575   &  & 496.723  &   & 0.929  &  & 0.613  \\ 
\midrule
	\multirow{3}{*}{$k$-fold CV} & Averaged & \multirow{3}{*}{1.821} & 1.514 & \multirow{3}{*}{0.987} & 0.901 & \multirow{3}{*}{1.061} & 0.726 & \multirow{3}{*}{0.585} & 0.607 \\ 
     & Best &      & 1.492 & & 0.905 & & 0.747 & & 0.601 \\    
     & Retrained & & 1.486 & & 0.901 & & 0.735 & & 0.604 \\ 
	[3pt]\multicolumn{1}{r}{\multirow{3}{*}{V\_}} & Averaged & \multirow{3}{*}{2.008} & 1.994 & \multirow{3}{*}{0.988} & 0.902 & \multirow{3}{*}{1.063} & \textbf{0.709} & \multirow{3}{*}{0.590} & 0.593 \\ 
     & Best &      & 2.020 & & 0.891 & & 0.830 & & 0.596 \\    	
     & Retrained & & 2.001 & & 0.899 & & \textbf{0.714} & & 0.587 \\ 
	[3pt]\multicolumn{1}{r}{\multirow{3}{*}{\_V}} & Averaged & \multirow{3}{*}{2.027} & 1.992 & \multirow{3}{*}{1.000} & 0.914 & \multirow{3}{*}{1.089} & 0.737 & \multirow{3}{*}{0.568} & \textbf{0.585} \\ 
     & Best &      & 1.637 & & 0.905 & & 0.785 & & 0.601 \\    
     & Retrained & & 2.001 & & 0.912 & & 0.763 & & \textbf{0.584} \\ 
	[3pt]\multicolumn{1}{r}{\multirow{3}{*}{\_V\_}} & Averaged & \multirow{3}{*}{2.036} & 1.967 & \multirow{3}{*}{0.997} & 0.896 & \multirow{3}{*}{1.083} & 0.792 & \multirow{3}{*}{0.574} & 0.588 \\ 
     & Best &      & 2.016 & & 0.899 & & 0.817 & & 0.589 \\    
     & Retrained & & 2.001 & & 0.891 & & 0.788 & & \textbf{0.584} \\ 
\midrule 
	\multirow{3}{*}{$k$-step AV} & Averaged & \multirow{3}{*}{2.166} & \textbf{1.272} & \multirow{3}{*}{1.016} & 0.848 & \multirow{3}{*}{0.731} & 0.854 & \multirow{3}{*}{0.636} & 0.680 \\ 
     & Best      & & \textbf{1.197} & & \textbf{0.816} & & 0.924 & & 0.620 \\ 
     & Retrained & & 1.873 & & 0.870 & & 0.765 & & 0.622 \\ 
	[3pt]\multicolumn{1}{r}{\multirow{3}{*}{\_V}} & Averaged & \multirow{3}{*}{2.262} & \textbf{0.693} & \multirow{3}{*}{1.036} & 0.855 & \multirow{3}{*}{0.724} & 0.830 & \multirow{3}{*}{0.640} & 0.616 \\ 
     & Best &      & \textbf{0.643} & & \textbf{0.821} & & 0.799 & & \textbf{0.566} \\    
     & Retrained & & 2.001 & & 0.881 & & \textbf{0.701} & & \textbf{0.566} \\ 
\midrule 
	\multirow{3}{*}{$k$-step FV} & Averaged & \multirow{3}{*}{2.218} & 1.591 & \multirow{3}{*}{1.035} & \textbf{0.839} & \multirow{3}{*}{0.733} & 0.742 & \multirow{3}{*}{0.648} & 0.629 \\ 
     & Best      & & 1.542 & & \textbf{0.814} & & 0.867 & & \textbf{0.573}\\ 
     & Retrained & & 1.695 & & 0.899 & & \textbf{0.712} & & 0.593 \\ 
	[3pt]\multicolumn{1}{r}{\multirow{3}{*}{\_V}} & Averaged & \multirow{3}{*}{2.379} & 1.639 & \multirow{3}{*}{1.023} & \textbf{0.840} & \multirow{3}{*}{0.702} & 0.805 & \multirow{3}{*}{0.636} & \textbf{0.573} \\ 
     & Best      & & 1.626 & & \textbf{0.818} & & 0.950 & & 0.693 \\ 
     & Retrained & & 1.752 & & 0.904 & & 0.738 & & 0.616 \\ 
\bottomrule
    \end{tabular*}
    \label{tab:generative_results}                            
\end{table}
\egroup

Achieving the stability of the outputs in generative mode is perhaps the most important issue, and we can immediately observe that all the cross-validation schemes successfully deal with it, compared to SV, where in almost half of the cases the outputs diverge in at least some trials (indicated by very large testing errors). This is a major advantage of using cross-validation. 

We can also observe, that SV consistently underestimates the validation error and is not able to choose the best hyper parameters for testing. The testing error is quite higher than validation, is never best and almost always quite far from that. 

In general SV is not a good validation strategy. However, choosing among the cross-validation schemes and variations is a bit more tricky.

We see that in three out of four cases, the $k$-step AV \_V scheme with either Best or Retrained final method gave the best testing errors, even averaging over five different random initializations. But there is not much consistency in which variation of $k$-step AV is best, and choosing a wrong one can hurt the performance quite much. $k$-step FV give similarly good results (the best with Gasoline), but has the same issue. 

$k$-fold CV, on the other hand, never gave the very best result, but for the longer and more stationary datasets of Sunspots and Electricity actually gave very good and stable results across all the variations. It can perhaps be recommended for these types of conditions as a reliable choice. 

The other two shorter and less stationary datasets Labor and Gasoline seem to benefit from forward-facing validations, with no peeking into the future like in $k$-fold CV. This can probably be explained by the generating processes likely having a one-directional ``drift'', thus the forward-facing validation schemes, that select models capable of predicting sequences following the training ones in time, win. Similar findings are concluded in \cite{CerqueiraEtAl2020}.

The Labor data set is the most difficult, as discussed above. Because it is so short there is a big variance in the results. SV performs relatively well here as it is trained on most of the scarce data, compared to the other forward-facing schemes AV and FV. Because the signal takes a sudden dive down in the testing interval it is hard to predict: almost all of the NRMSE errors come $> 1$, with the (lucky?) break-out exception of $k$-step AV \_V Best or Averaged.

On Gasoline, quite clearly the best results are produced with $k$-step AV or FW and Best final model, strongly suggesting that some data should be excluded from the training and/or validation. Looking at the signal in Figure \ref{fig:serieses} we can identify such suspect areas at the second half of it. Relatively good results with Averaged AV and FW, where the beginning of the signal is more emphasized corroborates this suspicion. 

It is hard to say from the results whether the gaps around the validation interval are beneficial or not. As expected, their use looks questionable with scarcer data, but overall there is not much consistency. Maybe they would be more beneficial in another testing (and application) setup. Also, using smaller gaps could be investigated as a possible compromise.

\begin{figure*}[htb]
\makebox[\linewidth][c]{%
\centering
\subfigure[SV Retrained]{\label{fig:a}\includegraphics[width=0.5\textwidth]{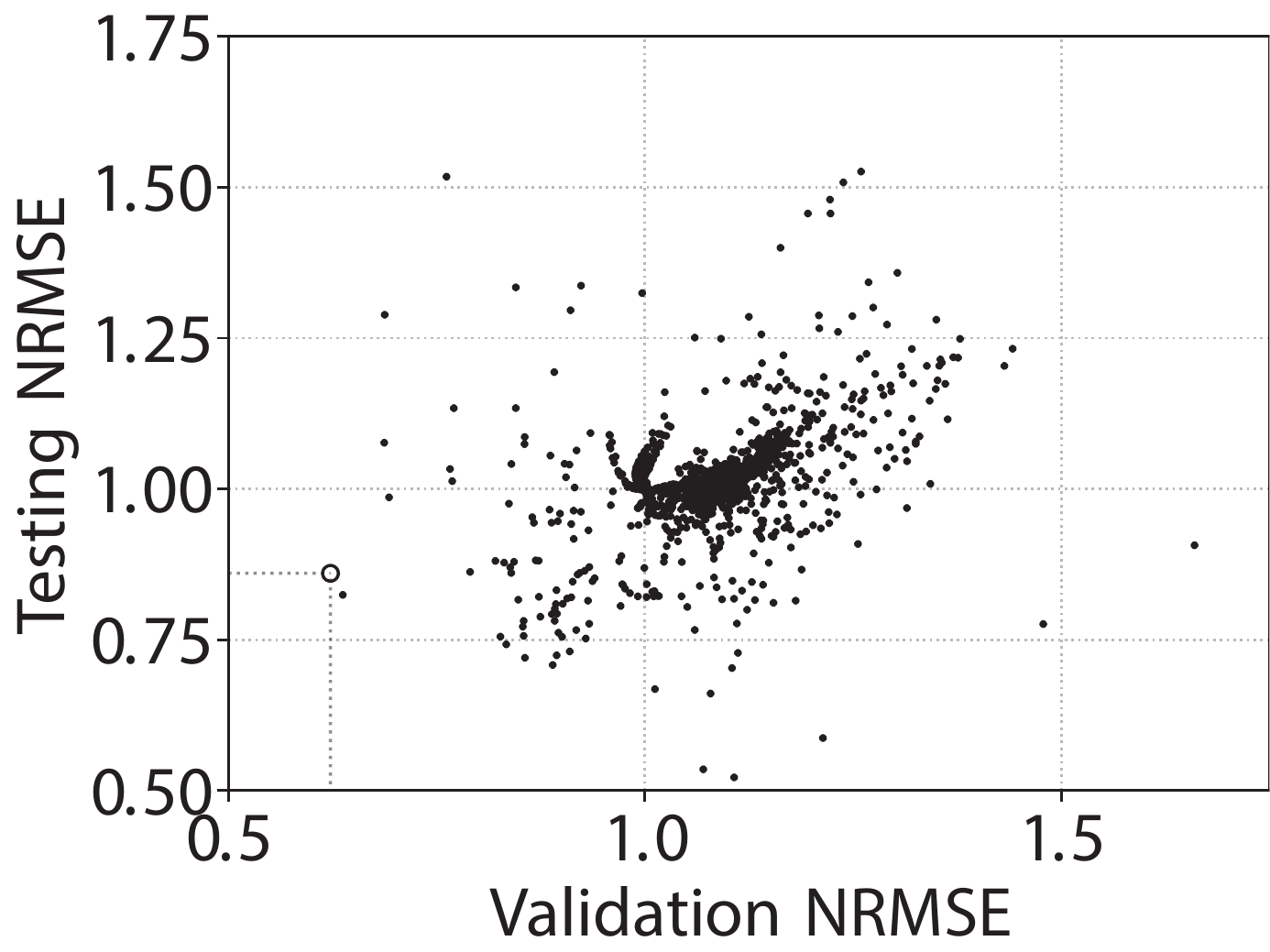}}%
\subfigure[$k$-step FV Averaged]{\label{fig:b}\includegraphics[width=0.5\textwidth]{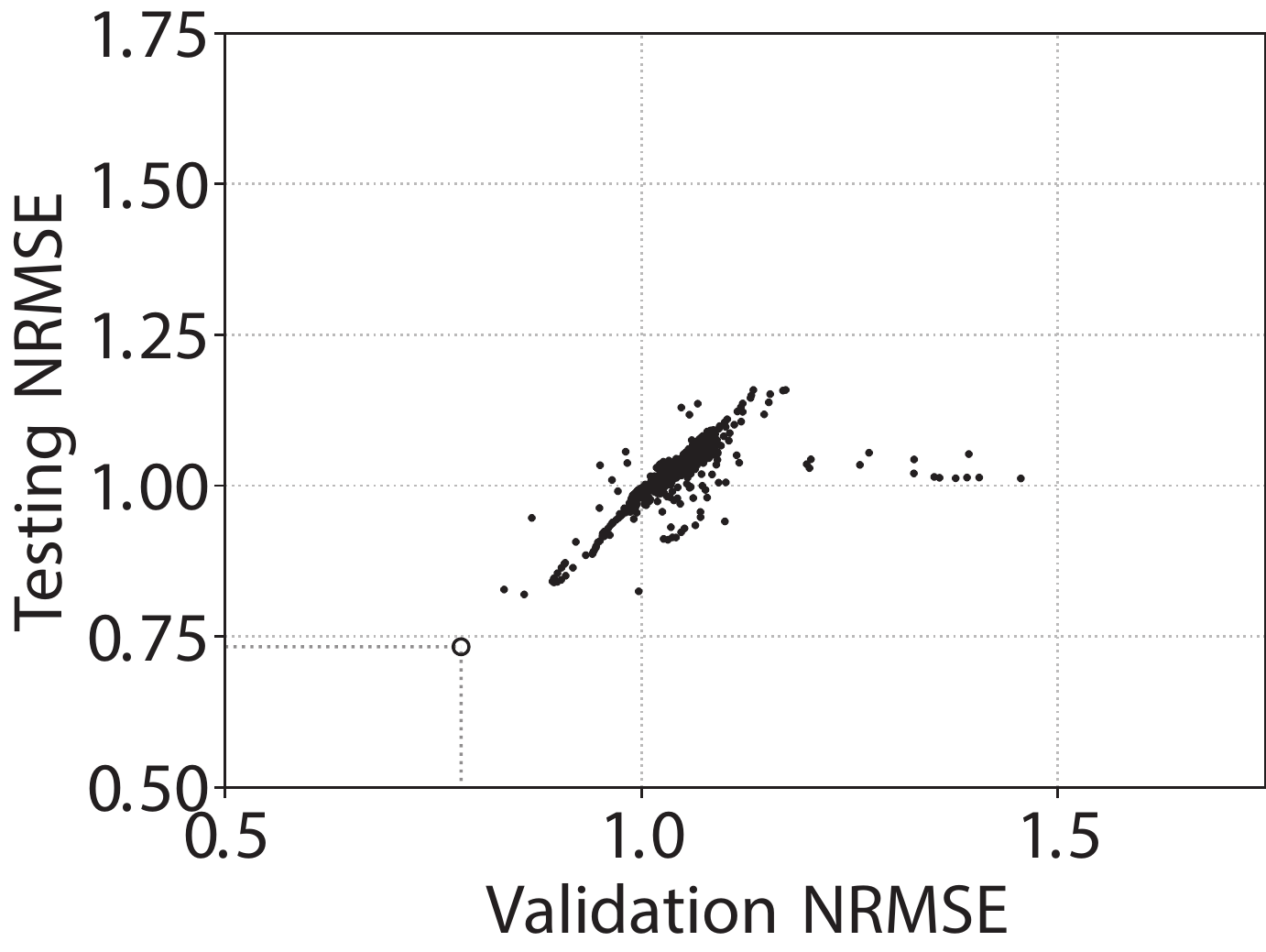}}%
}
\caption{Results of grid search on Electricity dataset (single run). Every point corresponds to one combination of hyper-parameters.}
\label{fig:val_test}
\end{figure*}

The validation vs. testing errors of SV and $k$-step FV validation schemes on the Electricity dataset with different hyper-parameter sets are presented in Figure \ref{fig:val_test}. We can see that while there exist outliers with very good testing performance in Figure \ref{fig:a}, they would never be selected by the validation. %
Their big difference between the validation and testing errors indicates that the lucky outliers were particular to the testing data and most likely would not do well on other. But most importantly, $k$-step FV gives a much better overall correlation between the validation and testing errors and a much better hyper-parameter set is picked based on validation in Figure \ref{fig:b} (the small circles).

\FloatBarrier 
\subsection{Time Series Output}\label{timeSeriesExperiments}

\begin{figure}[h]
 \centering
 \includegraphics[scale=0.685]{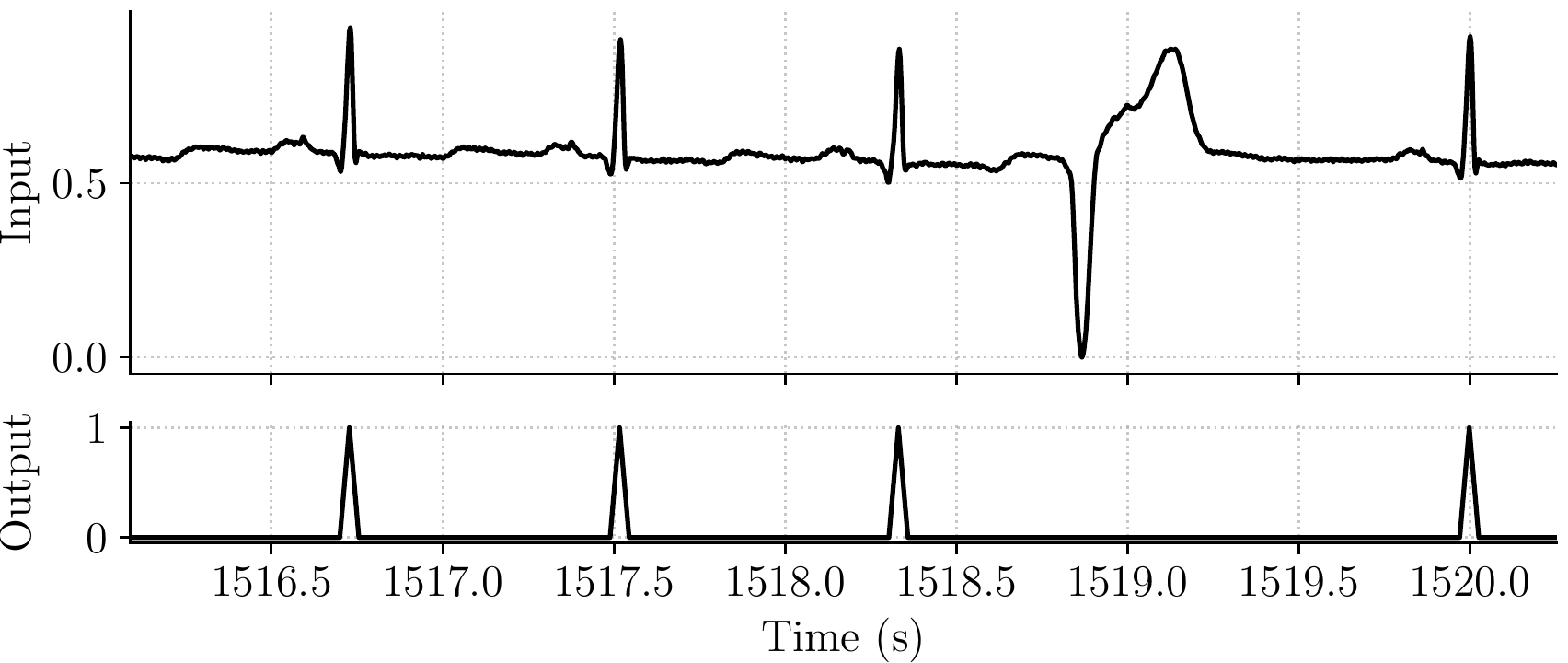}
\caption{A sample of the training data of the MIT-BIH dataset. }
 \label{fig:output_task}
\end{figure}

For a time series output task, we use a large dataset from MIT-BIH Arrhythmia Database \cite{Moody2001}\footnote{Publicly available at \url{https://www.physionet.org/physiobank/database/mitdb/}}. We take the first channel of the ambulatory ECG of the first patient (\#100), which consists of 650\,000 data points (about half an hour of ECG recording at 360\,Hz) as our input $\u(n)$ and heart beat annotations signal as the target output $\yt(n)$ for the model. We only annotate normal beats, which constitute the majority of all annotations. The annotation signal at the time step where the R peak of the normal heart beat occurs is 1 and fades linearly to 0 in the neighborhood of 5 time steps in both directions. A sample of the data is presented in Figure \ref{fig:output_task}. Note that only normal beats are to be recognized and, e.g., premature ventricular contraction seen here is not reflected in the target output indicator signal.

We use 70\,\% of the data for training and validation, and the rest for testing. The first 1.09\,\% of the training data (5\,000 time steps) were used for initialization and the rest was split for the different cross-validation schemes. We use $k=10$ folds and a validation fold size of 10\,\% in every scheme. For $k$-fold CV the size of each fold is 10\,\%, while for the $k$-step AV and FV, we used min. size of 50\,\% and the step size of 5\,\%. Finally, for SV we dedicate the last 10\,\% of the training data for validation and 90\,\% for the training. The same grid-search setup was used as described in Section \ref{generativeExperiments}, except that $\alpha \in \{ 0.1, 0.4, 0.7, 1\}$ and $\rho \in \{0.1, 0.25, 0.4, \ldots, 1.45 \}$. The mean validation NRMSE (variance was small) and testing error statistics for three types of final models and five runs are presented in Table \ref{tab:output_results}, together with illustratory best hyper-parameters in one of the runs. 

\bgroup
\setlength\tabcolsep{5.8pt}
\begin{table}[H]                           
 \centering
    \caption{NRMSE statistics of validation and testing on time series output task (mean $\pm$ standard deviation over five runs) with three types of final models, together with a sample of best hyper-parameters.}
    \begin{tabular}{l c c c c c c c }
    \toprule
         \multicolumn{2}{c} {\tableHeader{Validation}} & \multicolumn{3}{c} {\tableHeader{Testing error}} & \multicolumn{3}{c}{\tableHeader{Hyper-params}}    
     \\
     \cmidrule(r){1-2} \cmidrule(lr){3-5} \cmidrule(l){6-8}   
     \tableHeader{Scheme} & \tableHeader{Error} &  \tableHeader{Averaged} & \tableHeader{Retrained} & \tableHeader{Best} & $\alpha$ & $\rho$ & $\beta$ \\
         \midrule 
     SV  &  0.256  & {0.244 $\pm$ 0.033} & 0.242 $\pm$ 0.033  & {0.244 $\pm$ 0.033} & 0.1 & 1.45 & $0$ \\ 
     $k$-fold CV & 0.186  & 0.218 $\pm$ 0.004 &  0.217 $\pm$ 0.005  &  0.218 $\pm$ 0.005  & 0.4 & 0.85 & $10^{-9}$ \\ 
     $k$-step AV & 0.200  &  0.218 $\pm$ 0.008  &  0.216 $\pm$ 0.006  &  0.216 $\pm$ 0.008   & 0.7  & 0.85 & $10^{-9}$ \\ 
     $k$-step FV & 0.198  & 0.216 $\pm$ 0.005  & 0.217 $\pm$ 0.004 &  0.216 $\pm$ 0.004  & 0.4 & 0.85 & 0 \\ 
        \bottomrule
    \end{tabular}
    \label{tab:output_results}                            
\end{table}
\egroup

We see in Table \ref{tab:output_results} that in all the experiments the classic single validation SV produced the worst testing errors. ``Averaged'' and ``Best'' is the same in this case, since we only have one fold. Notice that SV had a very different validation error, picked a very different set of hyper-parameters, and has a high variance of the testing errors. On the other hand, all the cross-validation (and final model production) schemes gave very similar, consistent, and better results.

This indicates a classical situation where validation on a single data interval is not reliable and stable enough. The interval likely contained some unusual artefacts. Since the data is abundant in this case, other factors do not matter much, as long as validation is averaged over several folds.

\subsection{Time Series Classification}\label{classificationExperiments}

To evaluate the validation schemes on a classification task we turn to the classical Japanese Vowels dataset\footnote{Publicly available at \url{https://archive.ics.uci.edu/ml/datasets/Japanese+Vowels}}. This benchmark comes as 270 training and 370 testing samples taken from nine male speakers, where each sample consists of varying length 12 LPC cepstrum coefficients. The goal of this task is to classify a speaker based on his pronunciation of the vowel /ae/. In the training set, there are 30 samples from each speaker, while in the testing set this number varies from 29 to 88.

Zero test set misclassifications have been already achieved with ESNs on this benchmark in \cite{JaegerEtAl07si}, so we turn our efforts here to achieving better results with less refined ways of using ESNs. Models that only use the last state vector $\x(n)$ for every speaker are reported to at best be able to achieve 8 test misclassifications in \cite{JaegerEtAl07si}. We replicate this model and set up a hyper-parameter grid search as described in Section \ref{generativeExperiments}. We run an $18$-fold CV with validation (fold) size of 15 instances. As the dataset is a permutable set of sequences (the order does not matter), we only test the classical SV and the standard $k$-fold CV. We run the experiments 500 times having different random initializations of the ESN weights $\W{in}$ and $\W{}$. For each of the 500 experiments we run grid search separately and report the aggregated results in Table \ref{tab:jap_grid_search}. 

We also try individual regularizations for each validation split. When doing cross-validation, each split is evaluated on each regularization degree candidate. $\W{out}_i$'s of each split $i$ with their individual best regularization degrees are averaged. We call this variation ``IReg Averaged''. In another variation, the model was retrained on the whole data using the average of the best regularization degrees. We call it ``IReg Retrained'' in Table \ref{tab:jap_grid_search}. 

\begin{table}[htb] 
    \centering
    \caption{Result statistics on Japanese Wovels classification task}
    \begin{tabular*}{\textwidth}{@{\extracolsep{\fill}} l l r r r}
    \toprule
     \tableHeader{Validation} & \tableHeader{Final} & \tableHeader{Validation error} & \tableHeader{Test error} & \tableHeader{Misclassifications} \\
    \midrule
     \multirow{2}{*}{SV} &  As is & \multirow{2}{*}{0.471 $\pm$ 0.049}  & 0.495 $\pm$ 0.024 & 5.608 $\pm$ 3.032 \\ 
      & Retrained &  & 0.492 $\pm$ 0.024 & 5.447 $\pm$ 3.087 \\ 
        \midrule
         \multirow{2}{*}{$k$-fold CV}&  Averaged  & \multirow{2}{*}{0.490 $\pm$ 0.009} & 0.468 $\pm$ 0.008 & 3.388 $\pm$ 1.404 \\
         & Retrained &  & 0.468 $\pm$ 0.008 & 3.404 $\pm$ 1.413 \\
        \midrule
         \multirow{2}{*}{$k$-fold CV IReg} & Averaged & \multirow{2}{*}{0.485 $\pm$ 0.009} & \textbf{0.466} $\pm$ 0.007 & 3.309 $\pm$ 1.304 \\
         {} & Retrained&  &  0.467 $\pm$ 0.007 & \textbf{3.200} $\pm$ 1.346 \\
        \bottomrule
    \end{tabular*}
    \label{tab:jap_grid_search}                                

\end{table}

Statistics of NRMSE errors and misclassifications are presented in Table \ref{tab:jap_grid_search}. We see that all the CV variations significantly outperform all the SV variations. They also give more consistent results in terms of standard deviation. This confirms the benefits of cross-validation once again.

We also see that the individual regularization ``IReg'' further slightly improves both validation and testing errors, as well as classification. This is not surprising, as this allows for more degrees of freedom and a better fine-tuning.

\section{Speed Experiment}\label{speed}

\begin{figure}[h]
 \centering
 \includegraphics[scale=0.27]{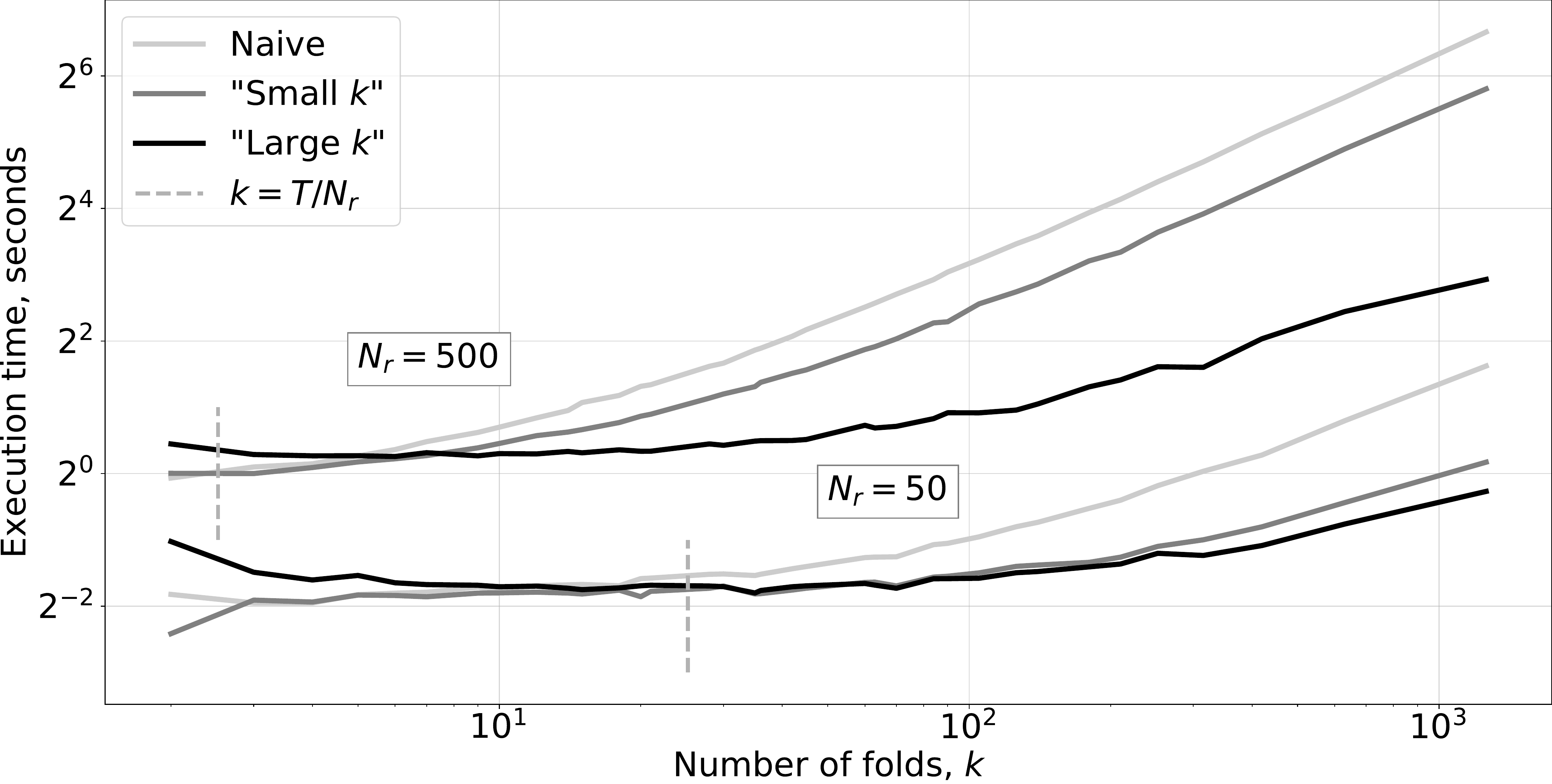}
\caption{Execution times of three cross-validation implementations versus the number of folds $k$ and reservoir size $\n{r}$. The results are shown in logarithmic scales.}
 \label{fig:time_measures}
\end{figure}

To test how well the theoretical time complexity reductions of our proposed methods translate into real world computational time savings, we conduct computational experiments. Since we are only investigating computational speed, we use a randomly generated data with $T = 1\,260$ and $\n{u} = \n{y} = 1$. We use two reservoir sizes $\n{r} = 50$ and $\n{r} = 500$, and run $k$-fold cross-validation with different number of folds $k$. The validation is done in a generative mode. The results are averaged over five runs. In these experiments we run a less space-efficient algorithm, that stores $\X$ and $\Yt$ directly with space complexity $\OO(N_r^2T)$, to gain some additional speed in all the methods. The experiments were run on an 2.0\,GHz 2 cores and 2 threads per core Intel Xeon CPU Google Colab machine. The results are presented in Figure \ref{fig:time_measures}. The upper three plots are for the reservoir size $\n{r} = 500$ and the lower three for $\n{r} = 50$.

We can see in Figure \ref{fig:time_measures} that the reservoir-optimized algorithm ``Small $k$'' is always performing better than the naive one, and ``Large $k$'' is slower for small $k$s, but performs better than both when the $k$ is large. There is also a significant difference in behavior depending on the size of the reservoir compared to $T$. It appears that for small reservoirs the reservoir optimisation of Section \ref{algorithm} is more important, while for large reservoirs the matrix inverse optimisation of Section \ref{largek} becomes more important. This makes sense: the effect of the reservoir optimization depends on how small $k$ is compared to $T/\n{r}$ and for small reservoirs $\n{r}$ this difference becomes large quickly, thus this optimization is more important. The matrix inverse optimization ``Large $k$'', on the other hand, gives biggest effect when $k$ is much larger than $T/\n{r}$, and thus is faster when $\n{r}$ is large.

\section{Conclusions}\label{discussion}

In this contribution we aimed at further improving the basic ESN (and overall RC) training aspects. ESNs are known and used for their fast training, but often rely on validation for good hyper-parameter tuning. We proposed more powerful validation schemes together with efficient algorithms to implement them, so that a better hyper-parameter tuning or model selection can be achieved with almost no sacrifice of the speed.

Cross-validation is still not that common in time series domain. Theoretical groundings for it are difficult and scarce, especially when lifting the assumption of stationary and searching for a model biased for a particular time interval of interest, but perhaps will be better developed in the future. For now we mostly rely on intuitions and empirical studies, which we also did here. 

We proposed and motivated different cross-validation schemes for RC models and also several options of how the final trained model can be produced. In addition to theoretical motivations, we tested the schemes on different types of real world tasks: time series output, prediction, classification. Our experiments show that typically cross-validation predicts testing errors more accurately and produces more robust results. It also can use scarce data more sparingly. There is no single winner among the cross-validation schemes. The results highly depend on the nature of the data. AV and FV validation schemes can apparently select for good ``forward-predicting'' models when the generating process is ``evolving'' and not quite stationary. The $k$-fold CV can use the data most sparingly and produce very stable and good results when the data are more stationary. The effectiveness of the different cross-validation schemes could in principle be used to judge about the (non-)stationarity of the analyzed data. Classical SV can also have its benefits, as it is the easiest to implement and can be seen as the last split of AV. As for using gaps around the validation fold, we have got mixed results in our testing setup, leading to no clear recommendation for or against. For stationary and ample, well-represented data the exact validation scheme might not matter much.

We have proposed two levels of RC cross-validation optimizations, eliminating the two biggest bottlenecks of the process, and making RC training almost as fast with cross-validation as it is without. The biggest bottleneck is running the reservoir with the data. Our algorithm allows running it up to three times, i.e., retain the original time complexity $\OO(\n{r}^2 T)$, irrespective of the number of folds $k$. This is a big saving for any reservoir computer. This optimization is enough to have roughly the same time complexity as with no cross-validation if $k$ is not very high.

The second bottleneck, that can come after that with an increase of $k$, is computing the matrix inverses to find all the output weights $\W{out}_i$ for each split $i$. We suggested the second level of optimization that makes this process fall back to the $\OO(\n{r}^2 T)$ time complexity, irrespective of the number of splits $k$. After these two bottlenecks are eliminated, the remaining therm that might become significant is the actual $k$ matrix multiplications to produce the $\W{out}_i$, which takes $\OO(k\n{r}^2\n{y})$ and depends on the dimension of the output. E.g., for a single-dimensional output our $k$-fold cross-validation algorithm has the same time complexity as the best implementation of a classical single validation, even if $k$ goes up to $T$ for a leave-one-out validation. 

All these time complexity optimizations are done with no increase in the best space complexity of a single-run training, which remains $\OO(\n{r}^2)$. 

This further sets apart the speed of ESN training from error backpropagation based recurrent neural network training methods, where cross-validation could also be used in principle but at a high computational cost.  

We have also run numerical simulations to empirically measure the computational time required to do the $k$-fold cross-validation with the different levels of optimization and other parameters. They confirm the computational savings and demonstrate which level of optimization is more pertinent in which situation. 

We share our code with the proposed methods at \\
\url{https://github.com/oshapio/Efficient-Cross-Validation-of-Echo-State-Networks}.

\section*{Compliance with Ethical Standards}

\textbf{Funding} This research was supported by the Research, Development and Innovation Fund of Kaunas University of Technology (grant No. PP-91K/19).

\textbf{Conflict of Interest} The authors declare that they have no conflict of interest.

\textbf{Ethical approval} This article does not contain any studies with human participants or animals performed by any of the authors.

\bibliographystyle{unsrt}     
\bibliography{ML2}

\end{document}